\documentclass{article} 
\usepackage{iclr2026_conference,times}

\usepackage{amsmath,amsfonts,bm}

\def\eqref#1{equation~\ref{#1}}

\def\1{\bm{1}}

\DeclareMathAlphabet{\mathsfit}{\encodingdefault}{\sfdefault}{m}{sl}
\SetMathAlphabet{\mathsfit}{bold}{\encodingdefault}{\sfdefault}{bx}{n}

\usepackage{hyperref}
\usepackage{url}
\usepackage{graphicx}
\usepackage{amsmath}
\usepackage{booktabs}
\usepackage{bm}
\usepackage{multirow}
\usepackage{wrapfig}
\usepackage{bbding}
\usepackage{pifont}

\title{Chronological Thinking in Full-Duplex \\ Spoken Dialogue Language Models}

\author{Donghang Wu$^{1,2,*}$ \ Haoyang Zhang$^{1,2,}$\thanks{Equal contribution} \quad  Chen Chen$^{1,\dag}$ \ Tianyu Zhang$^{3}$
\ \textbf{Fei Tian$^{2}$} \\ \textbf{Xuerui Yang$^{2}$ \ Gang Yu$^{2}$} \ \textbf{Hexin Liu$^{1}$\ \ Nana Hou$^{1}$} \ \ \textbf{Yuchen Hu$^{1}$ \ \ Eng Siong Chng$^{1,}$\thanks{\small{Corresponding authors: \texttt{chen1436@e.ntu.edu.sg; ASESChng@ntu.edu.sg} } } } \\
$^1$Nanyang Technological University \quad $^2$StepFun \quad $^3$Mila  
}

\iclrfinalcopy
\begin{document}

\maketitle

\begin{abstract}
Recent advances in spoken dialogue language models (SDLMs) reflect growing interest in shifting from turn-based to full-duplex systems, where the models continuously perceive user speech streams while generating responses. This simultaneous listening and speaking design enables real-time interaction and the agent can handle dynamic conversational behaviors like user barge-in. However, during the listening phase, existing systems keep the agent idle by repeatedly predicting the silence token, which departs from human behavior: we usually engage in lightweight thinking during conversation rather than remaining absent-minded. Inspired by this, we propose \textit{Chronological Thinking}, an on-the-fly conversational thinking mechanism that aims to improve response quality in full-duplex SDLMs. Specifically, chronological thinking presents a paradigm shift from conventional LLM thinking approaches, such as Chain-of-Thought, purpose-built for streaming acoustic input. (1) \textit{Strictly causal}: the agent reasons incrementally while listening, updating internal hypotheses only from past audio with no lookahead. (2) \textit{No additional latency}: reasoning is amortized during the listening window; once the user stops speaking, the agent halts thinking and begins speaking without further delay. Experiments demonstrate the effectiveness of chronological thinking through both objective metrics and human evaluations show consistent improvements in response quality. Furthermore, chronological thinking robustly handles conversational dynamics and attains competitive performance on full‑duplex interaction metrics.

\end{abstract}

\section{Introduction}
Speech is a natural and fundamental modality for human–computer interaction, offering intuitive, efficient, and expressive communication~\citep{slm_survey1, stepaudio}. Reflecting this importance, spoken dialogue language models (SDLMs) have become increasingly central in AI as advanced systems seek to support natural interaction. In academia, SDLMs remain an active area of research~\citep{gen_sdlm, salmduplex, kimi}, with a growing emphasis on end-to-end speech-to-speech dialogue systems that integrate speech understanding and generation within a unified interactive loop. \par

\begin{figure}[t]
\begin{center}
\begin{minipage}[b]{0.89\linewidth}
  \centering
  \centerline{\includegraphics[width=14cm]{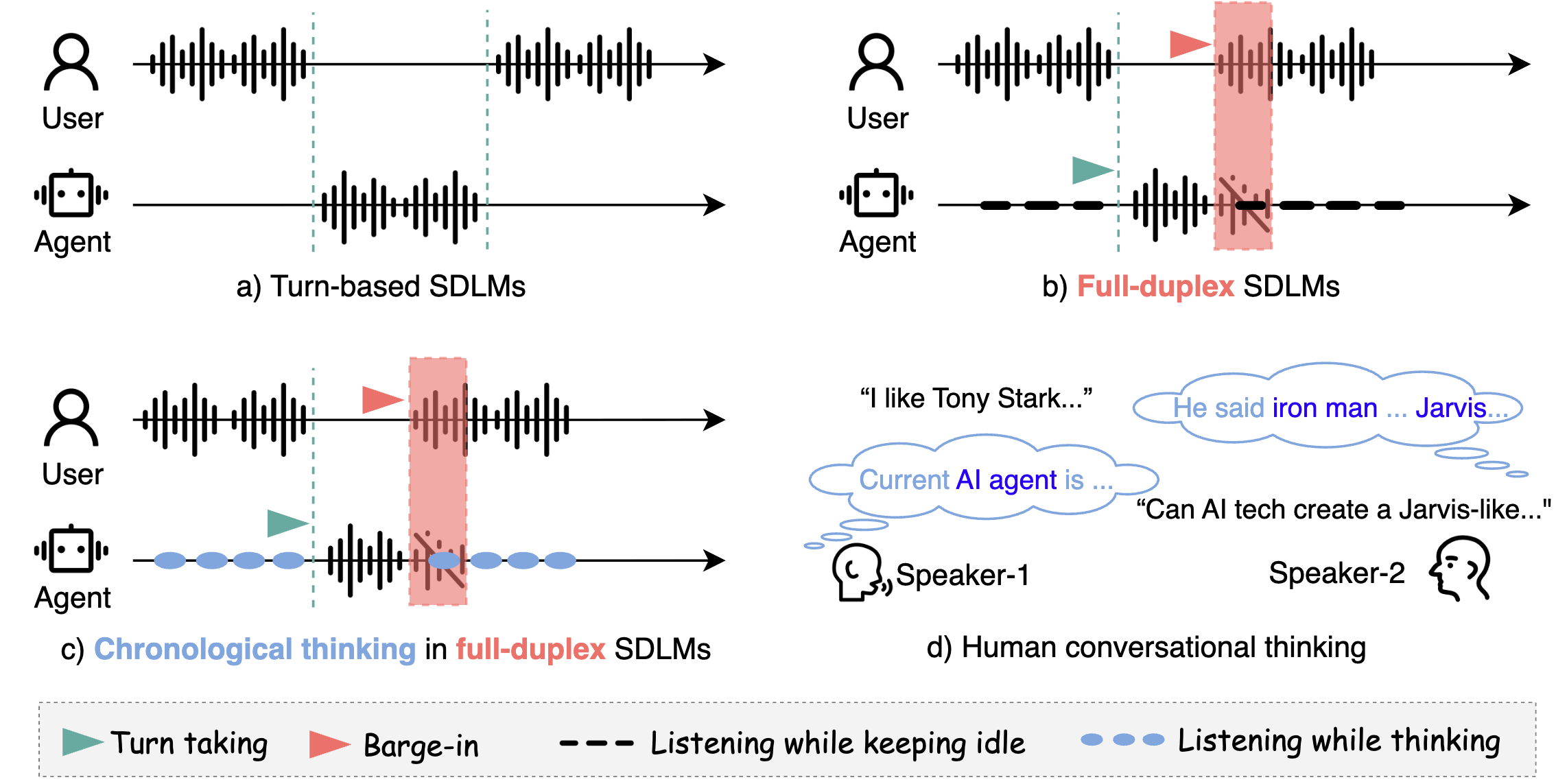}}
\end{minipage}
\vspace{-0.2cm}
\caption{Comparison of a) turn-based SDLMs, b) full-duplex SDLMs c) chronological thinking in full-duplex SDLMs (ours) and d) human conversational thinking patterns.}
\vspace{-0.5cm}
\label{fig:chronologicalthinking}
\end{center}
\end{figure}

More recently, full‑duplex models have garnered significant attention as a novel SDLM architecture~\citep{moshi, minmo, fireredchat}, departing from traditional turn‑based interaction by removing rigid listen‑then‑speak alternation~\citep{turn_based, Duplex_conversation, FlexDuo}, as illustrated in Figure~\ref{fig:chronologicalthinking}. In a full‑duplex system, the model continually ingests streaming user speech while synthesizing the corresponding response in real time. This ``always-on" agent delivers more natural, fluid, and human-like conversations, with the ability to proactively take turns, offer backchannel responses, make timely corrections, and gracefully yield with user barges-in~\citep{moshi}. \par


However, despite active exploration of full‑duplex models, we identify a common issue across existing designs: the agent is kept idle during user speaking by repeatedly predicting a ``silence token". This practice is problematic for two reasons: (1) in autoregressive models, prolonged repetition of a single token can be harmful, biasing the next‑token distribution and reinforcing degeneracy~\citep{Penalty_Decoding, learn_bias_repetition, loop}; and (2) it leaves the listening window underutilized, missing opportunities to form intent hypotheses and organize the forthcoming response. In contrast, human listeners perform lightweight, conversational thinking while listening—continually updating beliefs about the speaker’s intent and sketching response structure, as shown in Figure \ref{fig:chronologicalthinking}. This observation raises a central research question: \textit{Can such on-the-fly thinking be feasible within full‑duplex SDLMs?}


Notwithstanding the above, implementing such a mechanism under full‑duplex constraints is particularly challenging. First, streaming user speech imposes strict causality: the agent must reason incrementally while listening, updating hypotheses only from past audio without lookahead or access to a complete utterance~\citep{turn_based}. Second, since user speech can end at any moment, the reasoning process must be preemptible and amortized during listening; once the user stops, the agent should transition to speaking immediately without incurring additional latency~\citep{Incremental_Processing}. However, existing “thinking” techniques in LLMs, such as Chain-of-Thought (CoT)~\citep{cot}, are typically lengthy and post hoc, and therefore not directly applicable to the full-duplex setting. These limitations motivates the development of a new paradigm.
 \par
 
To address these challenges, we propose CT-SDLM, a full-duplex \textbf{SDLM} with a \textbf{C}hronological \textbf{T}hinking mechanism. Inspired by the Adaptive Control of Thought-Rational (ACT-R) theoretical framework, we propose different node types corresponding to specific modules in the typical ACT-R architecture~\citep{act-r}, which replace the repeated silence tokens in conventional full-duplex system. Given the streaming user speech input, these nodes are chronologically predicted based on real-time semantic segments, ensuring causality in the thinking process. This design is consistent with human conversational behavior, where we tend to form associations and generate responses incrementally based on the semantic fragments of the interlocutor’s speech, as shown in Figure~\ref{fig:chronologicalthinking}(d). Furthermore, compared with typical long and ad hoc language chain, the structured node representation significantly reduces the token cost of reasoning while retaining useful information in auto-regressive generation, even if the user stops speaking abruptly. Therefore, this compact thinking chain is preemptible, allowing the system to seamlessly switch to response generation without incurring any additional latency. We conduct both objective and subjective evaluations to verify the effectiveness of chronological thinking in full-duplex SDLMs. Across task-oriented dialogue~\citep{spokenwoz,urobench} and open-domain spoken QA~\citep{llamaq, webquestions} benchmarks, CT-SDLM consistently outperforms strong baselines in both A/B tests and quantitative metrics. In addition, evaluations on full-duplex interaction metrics confirm that CT-SDLM introduces no additional latency in turn taking and user barge-in, demonstrating its robustness to conversational dynamics. \par
Our contributions are summarized as follows: (1) We propose a full-duplex SDLM with chronological thinking---a strictly causal, on-the-fly reasoning mechanism that enables the model to incrementally process semantic segments during user speech. (2) Our design yields a compact, preemptible thinking process that replaces redundant silence tokens without adding latency, and achieves consistent gains over baselines in both subjective and objective evaluations. (3) We demonstrate that chronological thinking serves as a viable new paradigm for full-duplex interaction, with the potential to influence future directions in real-time dialogue modeling and human-machine communication. 



\section{Related work}
\label{RW}

\noindent\textbf{Full-Duplex Spoken Dialogue Systems.}
Early spoken dialogue systems involved turn-based architectures~\citep{sarikaya2002turn}, where user speech input and system output occurred sequentially. Recent advances have shifted toward full-duplex dialogue systems~\citep{veluri2024beyond,wang2024full,listen_speaking}, enabling the agent to simultaneously listen and speak. Research in this area has largely focused on engineering challenges such as streaming ASR~\citep{streaming_asr, he2019streaming, vad_analysis}, incremental TTS~\citep{Incremental_Processing, skerry2018towards}, and mechanisms for barge-in handling~\citep{rl_sdlm, schlangen2011general}. In addition, studies on incremental dialogue management~\citep{khouzaimi2016reinforcement, zhang2025llm} have explored how conversational agents can respond more naturally in overlapping speech conditions. Nevertheless, during the listening phase, existing systems primarily enforce silence by repeatedly predicting pause or silence tokens. This diverges from naturalistic human conversational behavior, where silent reasoning typically occurs.

\noindent\textbf{Reasoning in Language Models.}
A parallel line of work explores how language models perform reasoning through explicit intermediate steps. CoT~\citep{cot} and its extensions, such as Self-Consistency~\citep{selfconsistency} and Tree-of-Thoughts~\citep{tree_of_thought}, have shown that reasoning traces improve performance across arithmetic, logic, and commonsense reasoning. Recent methods such as Program-Aided Language Models (PAL)~\citep{pal} and Toolformer~\citep{toolformer} further highlight the benefits of externalized or structured reasoning. However, these approaches are designed for static text inputs, assuming access to the full problem before reasoning begins. They often rely on non-causal computation with hypothesis revision, which is incompatible with streaming conversational input.

\noindent\textbf{Incremental and Streaming Reasoning.}
Another relevant line of work investigates reasoning and generation under streaming or incremental input~\citep{calimeri2021stream}. In simultaneous machine translation, prefix-to-prefix frameworks~\citep{prefix-to-prefix} and monotonic attention models~\citep{Monotonic, Monotonic2} have been developed to balance accuracy and latency. Similar ideas in incremental decoding~\citep{incremental_decoding} allow models to generate partial outputs while processing incomplete inputs. In the LLM era, recent approaches such as StreamingLLM~\citep{StreamingLLM}, Medusa decoding~\citep{Medusa}, and attention sink methods~\citep{StreamingLLM} have examined how large models can operate efficiently under bounded memory and real-time constraints. These studies illustrate the feasibility of causal reasoning under partial input but rarely consider spoken dialogue. Our work addresses this gap by introducing chronological thinking, a causal reasoning mechanism tailored for full-duplex spoken dialogue, enabling models to think continuously while listening without delaying the onset of response generation.


\section{Method}
\label{method}
In this section, we introduce the proposed full-duplex SDLM with chronological thinking mechanism. We first describe the overall network architecture, which enables simultaneous input processing and response generation, and then detail the chronological thinking mechanism that replaces redundant silence during periods when the agent listens to the user. This enhancement aims to improve the model's intelligence, enabling it to possess human-like ability of thinking while listening.

\begin{figure}[t]
\begin{center}
\begin{minipage}[b]{0.9\linewidth}
  \centering
  \centerline{\includegraphics[width=13cm]{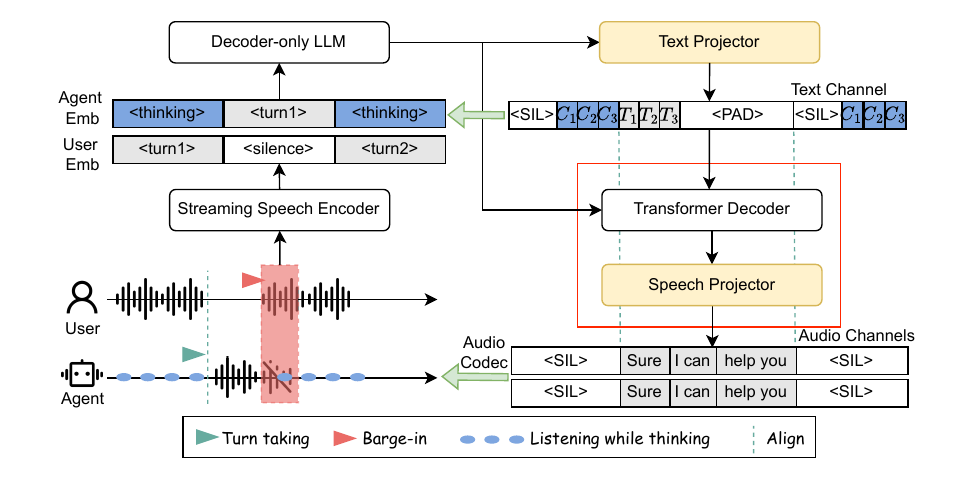}}
\end{minipage}
\caption{The network architecture of CT-Duplex with \textbf{chronological thinking} mechanism. Our model consists of a streaming speech encoder, a speech codec, an LLM backbone and a Transformer decoder. Compared to existing full-duplex SDLMs, chronological thinking is introduced during the listening phase.}
\label{fig:modelstructue}
\end{center}
\end{figure}

\subsection{Model architecture}
\label{model}
The network architecture of this paper is illustrated in Figure \ref{fig:modelstructue}. It accepts two input streams: the user speech stream and the agent speech and text stream. The user's speech stream is input into a streaming speech encoder operating at a frame rate of $12.5$Hz, producing continuous embeddings $\bm{X}\in \mathbb{R}^{T}$, where $T$ denotes the number of frames. These embeddings are projected by a modality adapter and then summed with the embeddings of agent text tokens before being fed into an LLM backbone.
To reduce the prediction burden on the LLM, instead of generating both the text and speech tokens with LLM \citep{salmduplex}, we only set the agent's text tokens $\bm{Y}^{\mathrm{txt}}$ as the prediction target for the LLM. This autoregressive process can be formulated as:
\vspace{-5pt}
\begin{equation}
\small
\begin{split}
    P_{\theta_l}(\bm{Y}^{\mathrm{txt}} | \bm{X}) = \prod_{t=1}^{T}P_{\theta_l}(Y^{\mathrm{txt}}_t|\langle \bm{Y}^{\mathrm{txt}}_{1:t-1}, \bm{X}_{1:t-1}\rangle),
\end{split}
\end{equation}
where $\theta_{l}$ denotes the parameters of the LLM backbone. After the LLM backbone generating the agent's text tokens, an autoregressive Transformer decoder is employed to predict the agent's speech tokens $\bm{Y}^{\mathrm{spc}}$. The inputs to this Transformer decoder include the agent's text tokens $\bm{Y}^{\mathrm{txt}}$ and previously predicted agent speech tokens $\bm{Y}^{\mathrm{spc}}$, conditioned on the LLM's last hidden states $\bm{h}$, which can be formulated as:
{
\setlength{\abovedisplayskip}{5pt}
\setlength{\belowdisplayskip}{5pt}
\begin{equation}
\small
\begin{split}
    P_{\theta_d}(\bm{Y}^{\mathrm{spc}} | \langle \bm{X}, \bm{Y}^{\mathrm{txt}}, \bm{h}\rangle) = \prod_{t=1}^{T}P_{\theta_d}(Y^{\mathrm{spc}}_t|\langle \bm{Y}^{\mathrm{spc}}_{1:t-1}, \bm{Y}^{\mathrm{txt}}_{1:t-1}, \bm{h}_t\rangle),
\end{split}
\end{equation}
}where $\theta_d$ represents the parameters of the Transformer decoder and $\bm{h}_t$ is LLM's hidden state at time step $t$. We employ Nanocodec~\citep{nanocodec} with Finite Scalar Quantization~\citep{fsq} to obtain agent speech tokens, generating speech codes at 12.5Hz. We jointly finetune the streaming speech encoder, LLM backbone, and autoregressive Transformer decoder using a multi-channel next token prediction training objective~\citep{ntp}.

\subsection{Agent Text and Speech alignment}
Existing methods aligns the agent's text and speech at the token level, prepends the \texttt{<BOS>} token at the beginning of an agent's turn, appends the \texttt{<EOS>} token at the end of the agent turn, and uses text padding tokens \texttt{<PAD>} to pad the gap between text and speech tokens~\citep{moshi, salmduplex}. Furthermore, during the user's turn, most full-duplex models require the LLM to predict silence tokens \texttt{<SIL>} to fill the agent's stream. The expected agent's text tokens in the $i$-th turn can be represented as: 
\vspace{-3pt}
\begin{equation}
\small
\begin{split}
\bm{Y}^{\mathrm{txt}}=[&\mathtt{<SIL>}, \cdots, \mathtt{<SIL>}, \mathtt{<SIL>}, \cdots, \mathtt{<SIL>}, \\[-2pt]
&\mathtt{<BOS>}, R_{i,1}, R_{i,2}, \cdots, R_{i,T}, \mathtt{<PAD>}, \cdots, \mathtt{<PAD>}, \mathtt{<EOS>}]
\end{split}
\end{equation}
\vspace{-14pt}

\noindent where $R_{i,t}$ is the response token at time step $t$ in turn $i$. When the agent needs to listen to the user's speech, the LLM repeatedly generates silence tokens \texttt{<SIL>}. When the user finishes speaking and it becomes the agent's turn to respond, the LLM outputs a \texttt{<BOS>} token, marking the beginning of the agent's turn. When the agent's turn ends or is interrupted by the user, the LLM outputs an \texttt{<EOS>} token, after which it resumes repeatedly generating silence tokens. This mechanism of outputting silence segments to maintain a listening state is a common paradigm in end-to-end full-duplex SDLM systems~\citep{moshi, salmonn, salmduplex}. 

However, forcing an LLM to repeatedly predict the same silence token not only lives the listening window unexploited but also degrades its performance by biasing the next-token distribution~\citep{loop}. In fact, during human conversation, when listening to others speak, the brain does not simply remain idle; instead, it engages in thinking while processing the speaker's input. Similarly, a SDLM should not continuously output silence tokens during the user's turn. Instead, it should perform the thinking process as the user's input unfolds streamingly. 

\begin{table}[t]
\vspace{-5pt}
\caption{The definitions of the five node types in chronological thinking chains and their corresponding relationships with different modules in the ACT-R theory}
\label{tab:node}
\renewcommand\arraystretch{1.1}
\begin{center}
\begin{tabular}{lcc}
\toprule
\multicolumn{1}{c}{Node Type}  &\multicolumn{1}{c}{DESCRIPTION} &\multicolumn{1}{c}{ACT-R}
\\ \midrule
Entity         &Extracts entities from the dialogue.    &Visual module\\
Intent         & Represents the user's goal            &Goal module\\
Action         & Denotes the agent's executable operation &Manual module\\
Knowledge      & Retrieves factual or procedural knowledge. &Declarative module\\
Logic          & Captures rules or logic generated by the agent &Production system \\
\bottomrule
\end{tabular}
\end{center}
\vspace{-15pt}
\end{table}
\subsection{chronological thinking}
\label{thinkingchain}
A straightforward approach to streaming thinking is to use the streaming ASR transcript text as the thinking content \citep{streaming_asr, he2019streaming, vad_analysis}. We discuss the comparison between this method and the approach proposed in this paper in Appendix \ref{app:asr}. Inspired by research in human cognitive architecture, particularly the ACT-R theory, which divides human cognition into distinct modules, including the visual module for recognizing entities, the goal module for maintaining current intentions, the declarative module for retrieving knowledge, the manual module for controlling actions, and the production system for managing production rules~\citep{act-r}, we define five distinct types of nodes, which form a chain-structured chronological thinking content. The node types and their corresponding relationships with different modules in the ACT-R framework are shown in Table \ref{tab:node}.

The chronological thinking chain grows with the user's input. Each time a semantic segment from the user is received, the agent obtains one or more of the five types of nodes. These nodes could be any one of these five node types. There is no fixed order for the nodes. \textbf{The type of node generated depends solely on the semantics}. For example, if the user input is ``Help me order a restaurant to celebrate my birthday this weekend.", a chronological thinking chain would be:

\vspace{-3pt}
\begin{verbatim}
[Help me]{INTENT} Request assistance
[order a restaurant]{ACTION} Initiate booking process
[to celebrate my birthday]{LOGIC} Purpose: birthday celebration
[this weekend]{ENTITY} Timeframe: weekend
{KNOWLEDGE} Birthday: Decorations, discounts, or special perks
\end{verbatim}
\vspace{-3pt}

\noindent A complete dialogue containing chronological thinking chains is shown in Appendix \ref{example}. The transcribed speech content within [$\cdot$] is only used during the data generation phase to control the generation of chronological thinking chains and is not included in the input and output stream of the full-duplex SDLM. The content within \{$\cdot$\} represents the node type, followed by the node's attributes. For a full-duplex SDLM, a chain node is formatted as:
\vspace{-3pt}
\begin{verbatim}
{Node type} Node attributes
\end{verbatim}
\vspace{-3pt}

We employ the Qwen2.5-72B-Instruct LLM model to generate chronological thinking chains based on input dialogue data~\citep{qwen2.5}. 
After obtaining the chronological thinking chain, we convert it into tokens, denoted as $\bm{C}$, prepend the starting token \texttt{<BOC>} and append the ending token \texttt{<EOC>}, then put them to the positions originally occupied by silence tokens. Considering causality and latency requirements, we adopt the following strategies of varying lengths of $\bm{C}$:

We first define the length of chronological thinking chain tokens as $M$ and the original silence tokens's length as $S$. For cases where $M$ is less than or equal to $S-2$, we replace the last $M+2$ silence tokens with \texttt{<BOC>}, thinking chain tokens, and \texttt{<EOC>}. This ensures that the thinking tokens appear as late as possible, striving to ensure that the thinking tokens corresponding to a semantic segment appear later than the semantic segment in user's speech. Thus, the expected agent text tokens in the turn $i$ can be expressed as:
\vspace{-1pt}
\begin{equation}
\small
\begin{split}
\bm{Y}^{\mathrm{txt}}_i = [&\mathtt{<SIL>}, \cdots, \mathtt{<SIL>}, \mathtt{<BOC>}, C_{i,1}, C_{i,2}, \cdots, C_{i,M}, \mathtt{<EOC>}, \\[-3pt]
&\mathtt{<BOS>}, R_{i,1}, R_{i,2}, \cdots, R_{i,T}, \mathtt{<PAD>}, \cdots, \mathtt{<PAD>}, \mathtt{<EOS>}],
\end{split}
\end{equation}
where $C_{i,m}$ is the $m$-th chronological thinking chain token in the turn $i$, and the number of \texttt{<SIL>} equals to $S-(M+2)$.

When $M$ is greater than $S-2$, we first tokenize each chain node into tokens, and denote the number of tokens for each node as $M_1$, $M_2$, $\cdots$, $M_N$, where $N$ is the number of nodes. We then retain the first $n$ nodes such that $M'=\sum_{j=1}^{n}M_j\le S-2$, and $M'+M_{n+1}>S-2$. We replace the last $M'+2$ silence tokens with chronological thinking chain tokens formed by the first $n$ nodes, as well as \texttt{<BOC>} and \texttt{<EOC>}. Thus, the expected agent text token in the $i$-th turn can be written as:
\begin{equation}
\small
\begin{split}
\bm{Y}^{\mathrm{txt}}_i=[&\mathtt{<SIL>}, \cdots, \mathtt{<SIL>}, \mathtt{<BOC>}, C_{i,1}, C_{i,2}, \cdots, C_{i,M'}, \mathtt{<EOC>}\\
&\mathtt{<BOS>}, R_{i,1}, R_{i,2}, \cdots, R_{i,T}, \mathtt{<PAD>}, \cdots, \mathtt{<PAD>}, \mathtt{<EOS>}], 
\end{split}
\end{equation}
where the number of \texttt{<SIL>} equals to $S-(M'+2)$. We discuss the completeness of thinking chains in Appendix \ref{sec:completeness}. Finally, we use the text tokens with chronological thinking chains and the speech tokens to train the SDLM with multi-channel next token prediction~\citep{ntp}.

\section{Experiments}
\label{exp}

\subsection{Data generation}
\label{data}
Existing real-world conversational datasets, such as Fisher conversation dataset~\citep{fisher}, focus mainly on casual conversations, are insufficient to train the SDLM to respond to diverse human inquiries \citep{moshi, salmduplex, turntakingability}. To enhance the model's reasoning capabilities, we generate challenging dialogue data through synthetic methods. We first use seed content and an LLM to generate textual conversations, then convert these conversations into speech using a multi-speaker TTS system with voice cloning capabilities.
\begin{wraptable}{r}{0.45\textwidth}
\vspace{-15pt}
\caption{Hours of synthetic training data}
\label{tab:dataset}
\renewcommand\arraystretch{1.0}
\begin{center}
\begin{tabular}{lc}
\toprule
Dataset  & Hours \\
\midrule
GenConv         & 10.5k \\
SpokenWOZ-G    & 2.0k \\
Llamaq-G       & 2.7k \\
\bottomrule
\end{tabular}
\end{center}
\vspace{-20pt}
\end{wraptable}

To create data for general conversation, we first curate a wide range of topics from sources like Wikipedia, covering general knowledge, common sense, and current events. These topics serve as seeds for Qwen2.5-72B-Instruct LLM~\citep{qwen2.5} to generate the textual dialogues, formulating a dataset named \textit{GenConv}.

To train the model's ability in scenarios requiring reasoning, we introduce SpokenWOZ and select its training set as the seed dataset and create \textit{spokenWOZ-G}. SpokenWOZ encompasses various reasoning scenarios, including those requiring cross-turn information, temporal, mathematical, and semantic reasoning~\citep{spokenwoz}. We prompt Qwen2.5-72B-Instruct to generate topically-related dialogues with SpokenWOZ's format. We then calculate the similarity between the SpokenWOZ's dialogues and generated dialogues using the \textit{thefuzz} python library and discard any generated dialogue with a similarity score over 90\%. Similarly, to enhance the model's knowledge base, we create the \textit{Llamaq-G} dataset by applying the same generation scheme to generate dataset with format like LLama Questions ~\citep{llamaq}. 

After generating the textual dialogues and their corresponding chronological thinking chains, we synthesize the audio using Step-Audio-TTS-3B~\citep{stepaudio}, a large-scale text-to-speech model capable of high-quality voice cloning. To ensure speaker diversity, we build a speaker library by collecting and cleaning over 50,000 single-speaker audio clips from various sources such as short-form videos, streaming media, and podcasts to serve as the prompt audio. For each dialogue, two distinct speakers are randomly selected from this library, and their voices are cloned using Step-Audio-3B to produce a complete, natural-sounding multi-turn spoken conversation~\citep{stepaudio}. The total hours for each dataset are summarized in Table \ref{tab:dataset}.

We follow the method in~\citep{salmduplex} to create barge-in events: each dialogue turn has a random 50\% probability of cutting off the agent's speech to allow the user to barge in. When a barge-in occurs, a $0.64$s delay is enforced before the agent ceases speaking. We introduce a delay of 0.32s between the end of the user's speech and the start of the agent's response to enhance the naturalness of the dialogue. As demonstrated in~\citep{salmduplex}, this approach enables the model to effectively learn the barge-in behavior.

\subsection{Experimental settings}
\label{expset}
The model is implemented using the NeMo Toolkit~\citep{nemo} \footnote{https://github.com/NVIDIA-NeMo/NeMo/tree/main/nemo/collections/speechlm2} and trained on 8 L40s (48G) GPUs. The LLM backbone is initialized from the Qwen2.5-1.5B-Instruct~\citep{qwen2.5}. The speech encoder, text tokenizer and speech codec follow the ones in~\citep{salmduplex}. The optimizer is AdamW with an inverse Square Root Annealing learning rate schedule. The learning rate starts from 3e-4 with a warm-up of 2500 steps. We use Whisper-large-v3 to transcribe the generated speech into text for calculating evaluation metrics \citep{whisper}.

\subsection{Evaluation Data and Metrics}
We utilize SpokenWOZ to validate the model's response quality in scenarios requiring reasoning~\citep{spokenwoz}. Additionally, we employ the MtBenchEval from URO-Bench, a multi-turn dialogue evaluation dataset to assess the model's performance in daily conversations without complex reasoning \citep{urobench}. We employ the GPT scores generated by \texttt{gpt-4o-mini}, 
ranging from 0 to 5 to evaluate the performance. The prompts used is from URO-Bench \citep{urobench}. We also utilize the text BLEU score and Sentence-BERT similarity to evaluate the similarity between the generated responses and the target content \citep{bleu, sentence-bert}. 

To evaluate the model's factual knowledge capability, we introduce the Llama Questions and Web Questions datasets~\citep{webquestions,llamaq}. The metrics utilized is accuracy.

We follow~\citep{turntakingability} to evaluate the turn-taking and barge-in performance of the full-duplex SDLM. The metrics include:  
(1) Turn-taking latency: The delay in the agent's response to the user's query in the first dialogue turn;  
(2) Barge-in latency: The time between the user's interruption and the agent stopping speech;  
(3) Barge-in success rate: The percentage of cases where the agent stops speaking within 1.5s after the user interrupts;  
We employed the \textit{impatient} dataset in \citep{turntakingability}, where interruptions occur approximately every 2 seconds on average, to evaluate turn-taking and barge-in performance.

\begin{table}[t]
\vspace{-15pt}
\caption{Performance on SpokenWOZ and MtBenchEval in terms of GPT score, BLEU, and Sentence-BERT. GT-LM is an optimal cascaded system that feeds ground-truth user turns to the LLM. We use ``$thk$" to denote the proposed chronological thinking. The results of SALM-Duplex are reproduced by ourselves.}
\label{tab:gpt}
\renewcommand\arraystretch{1.0}
\setlength{\tabcolsep}{3pt}  

\begin{center}
\begin{tabular}{lccc|ccc}
\toprule
\multirow{2}{*}{Method}  &\multicolumn{3}{c|}{SpokenWOZ} &\multicolumn{3}{c}{MtBenchEval}\\ \cmidrule(lr){2-4} \cmidrule(lr){5-7} &{GPT score}
&{BLEU} &Sentence-BERT & {GPT score} 
&{BLEU} &Sentence-BERT
\\ \midrule
GT-LM & 2.48 & 7.60 & 0.55& 3.15& 10.18& 0.73
\\ \midrule
SALM-Duplex$^{*}$         & 2.11 & 8.73 & 0.34& 2.25 &5.31 & 0.47\\
CT-Duplex w/o $thk$   & 2.40 & 12.92 & 0.52& 2.39 &7.00 & 0.64\\
CT-Duplex w/ $thk$    & \textbf{2.61} &\textbf{16.30} &\textbf{0.59} & \textbf{2.44} & \textbf{7.34} &\textbf{0.67} \\
\bottomrule
\end{tabular}
\end{center}
\vspace{-15pt}
\end{table}
\begin{table}[t]
\vspace{-10pt}
\caption{Performance of different methods on Llama Questions and Web Questions benchmark in accuracy (\%). Results of baseline systems are taken from \citep{glm4voice}. The results of SALM-Duplex are reproduced by ourselves. We use ``$thk$" to denote the proposed chronological thinking.}
\label{tab:acc}
\renewcommand\arraystretch{1.0}
\begin{center}
\begin{tabular}{lccccc}
\toprule
{Method}  &{Modality} &{\# Params} &{Full-duplex} &{Llama Questions} &{Web Questions}
\\ \midrule
TWIST  & S$\rightarrow$S &7B& \ding{55} & 4.0 & 1.5\\
SpeechGPT  & S$\rightarrow$T & 7B& \ding{55}& 21.6 & 6.5\\
Spectron  & S$\rightarrow$T & 1B& \ding{55}& 21.9 & 6.1\\
Moshi  & S$\rightarrow$S & 7B& \ding{51}& 21.0 & 9.2\\
GLM-4-Voice  & S$\rightarrow$S & 9B& \ding{55}& 50.7 & 15.9\\
\midrule
SALM-Duplex$^{*}$  & S$\rightarrow$S & 1.5B& \ding{51}& 15.0 & 6.7\\
CT-Duplex w/o $thk$  & S$\rightarrow$S & 1.7B& \ding{51}& 30.4 & 13.2\\
CT-Duplex w/ $thk$   & S$\rightarrow$S& 1.7B & \ding{51} &\textbf{31.4} & \textbf{13.3} \\
\bottomrule
\end{tabular}
\end{center}
\vspace{-20pt}
\end{table}

\subsection{Results}
\label{results}
\subsubsection{Reasoning quality}
We evaluate the performance of the proposed CT-Duplex model with and without the chronological thinking mechanism, denoted as CT-Duplex w/ $thk$ and CT-Duplex w/o $thk$, respectively. For comparison, we also include the method from \citep{salmduplex}, named SALM-Duplex. The network architecture of SALM-Duplex is nearly identical to ours, with the key difference being that its LLM backbone simultaneously predicts both text tokens and audio tokens, whereas our LLM backbone only predicts text tokens, and an additional Transformer decoder is used to predict audio tokens. Furthermore, by feeding the LLM backbone with the ground-truth text of user inquiries and using the generated text to calculate scores, we establish an optimal cascaded system (GT-LM) to compare with our proposed method. The evaluation result of the reasoning abilities of different methods is shown in Table \ref{tab:gpt}. It can be observed that the integration of chronological thinking has enhanced response quality, \textbf{especially in scenarios requiring complex reasoning}, as evidenced by 8.75\% improvements on the SpokenWOZ benchmark. For everyday multi-turn dialogues evaluated on MtBenchEval, the observed gains are relatively modest (2.09\%). 

Meanwhile, Table \ref{tab:gpt} shows that the GT-LM method, which uses ground-truth user inquiry texts, performs worse than the proposed chronological thinking method on SpokenWOZ, but better than the method without thinking. However, on MtBenchEval, it achieves the best performance. This is because the output obtained by GT-LM is essentially the result of the text LLM without CoT. Therefore, for scenarios that require certain reasoning, this method performs worse than the thinking-enabled CT-Duplex w/ $thk$ method. For scenarios that do not require reasoning, this method achieves optimal results due to the ideal input.

When compared to SALM-Duplex, both CT-Duplex w/ and w/o $thk$ achieve better performance. This demonstrates the effectiveness of decoupling audio token prediction from the LLM backbone.

\subsubsection{Factual knowlege capability}
We further evaluate the model's level of factual knowledge. Table \ref{tab:acc} shows the accuracy of our proposed method compared to baseline methods on both Llama Questions and Web Questions. It can be observed that when it comes to benckmarks requiring factual knowledge, the proposed choronological thinking method showed negligible improvement. This is because factual knowledge-based Question-Answering tasks require minimal reasoning, as the model only needs to possess the relevant knowledge to answer questions correctly. Compared to SALM-Duplex, our models still achieve higher accuracy as we alleviate the predictive burden on the LLM. Meanwhile, when compared with other baseline methods, it can be observed that except for GLM-4-Voice, our models outperform all baseline approaches, even though baseline methods such as Moshi have significantly more parameters than our models (7B vs. 1.7B). Although the GLM-4-Voice method achieves much higher accuracy than our proposed models, its 9B-parameter count far exceeds that of our models. Additionally, its half-duplex structure, which is not constrained by real-time and causal requirements, also contributes to its higher accuracy. In summary, the full-duplex SDLM proposed in this paper demonstrates strong factual knowledge retention capabilities.

\begin{table}[t]
\vspace{-10pt}
\caption{Evaluation of conversational behaviors of different methods on the \textit{Impatient} dataset proposed in \citep{turntakingability}, in terms of turn-taking latency, barge-in latency, and barge-in success rate. Results of baseline systems are taken from \citep{turntakingability}. We use ``$thk$" to denote the proposed chronological thinking}
\label{tab:tt}
\renewcommand\arraystretch{1.0}
\begin{center}
\begin{tabular}{lcc|cc}
\toprule
\multirow{2}{*}{Method}  &\multirow{2}{*}{E2E} &{Turn-taking}  &\multicolumn{2}{c}{Barge-in}\\ \cmidrule(lr){3-5} & & {Latency ($\downarrow$)} & {Latency ($\downarrow$)} & {Success rate ($\uparrow$)} 
\\ \midrule
Freeze-Omni & \ding{55}        & 1.17 & 1.20 & 79.50\% \\
dGSLM  &\ding{51}       & 0.57 & 0.86 & 85.00\% \\
Moshi &\ding{51}         & n.a. & 0.81 & 55.10\% \\
ORISE &\ding{51}    & 0.43  & 0.61 & 96.80\% \\
\midrule
SALM-Duplex$^{*}$ & \ding{51}        & {0.92}&  {0.69} & 87.50\%\\ 
CT-Duplex w/o $thk$ & \ding{51}        & \textbf{0.45}& \textbf{0.53} & 88.63\% 
\\
CT-Duplex w/ $thk$ & \ding{51}        & 0.68 & 0.54 & \textbf{94.05\%} 
\\
\bottomrule
\end{tabular}
\end{center}
\vspace{-15pt}
\end{table}

\subsubsection{Turn-taking and barge-in evaluation}
Table \ref{tab:tt} presents a comparison of the turn-taking and barge-in performance between the proposed method and the baseline method. It can be observed that the results are comparable with and without the thinking mechanism., whether in terms of turn-taking latency or barge-in behavior. Although the CT-Duplex w/o $thk$ method achieves a lower turn ($0.20$s lower), this difference has a negligible impact on the user experience in dialogue systems. 
Furthermore, the CT-Duplex w/ $thk$ method achieves a higher barge-in success rate. 
These results demonstrate that chronological thinking introduced in this work does not impair the full-duplex SDLM's turn-taking or barge-in abilities. That is because the proposed method generates nodes of the thinking chain chronologically with the input, introducing no additional computational overhead or extra latency. Besides, we only replace the silence tokens in the original full-duplex SDLM with thinking chain tokens, ensuring that the thinking process occurs exclusively during the listening phase without altering any response tokens of the SDLM.

\begin{wrapfigure}
{r}{0.45\textwidth}
\vspace{-50pt}
\renewcommand\arraystretch{1.1}
\begin{center}
\begin{minipage}[b]{0.9\linewidth}
  \centering
  \centerline{\includegraphics[width=7cm]{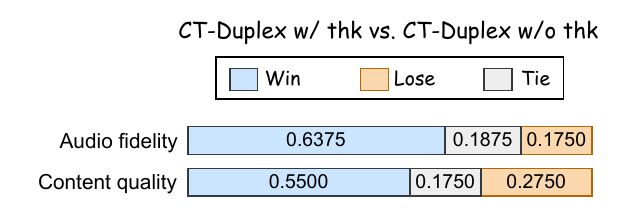}}
\end{minipage}
\caption{The A/B test results, including evaluations of both audio generation quality and response content quality.}
\label{fig:abtest}
\end{center}
\vspace{-10pt}
\end{wrapfigure}

\subsubsection{Subjective results}
We conduct an A/B test to validate the subjective evaluation results of the proposed chronological thinking method, aiming to assess its performance in terms of audio fidelity and response content quality\citep{abtest}. The experimental setups are detailed in Appendix \ref{sec:abtest}. The test dataset is sourced from SpokenWOZ. We select 10 fluent English speakers to evaluate the audio quality and content of the responses generated by CT-Duplex w/ $thk$ and CT-Duplex w/o $thk$. The experimental results are shown in Figure \ref{fig:abtest}. It can be observed that the subjective metrics for both audio quality and response content of CT-Duplex w/ $thk$ are superior to those of CT-Duplex w/o $thk$. Although the proposed thinking method is designed to improve response content quality, the enhancement in response quality leads to a lower loss in agent text prediction. This, in turn, allows the model to learn more from the audio prediction loss, thereby achieving a higher level of audio quality. 

\section{Conclusion}
\label{conclusion}
This paper proposes a chronological thinking mechanism that enables full-duplex SDLMs to possess a human-like thinking-while-listening ability during conversations. Inspired by research on human cognitive architecture, we introduce a chronological thinking chain comprising five distinct node types, each corresponding to components of the ACT-R framework. By replacing silence tokens in conventional full-duplex SDLMs with chronological thinking chain tokens, we achieve causal and no-additional-latency thinking during listening phases. Objective and subjective evaluation results demonstrate that the proposed method achieves higher response quality, especially in scenarios requiring reasoning, without compromising the turn-taking and barge-in performance of SDLMs.



\bibliography{iclr2026_conference}

@article{salmduplex,
  title={Efficient and Direct Duplex Modeling for Speech-to-Speech Language Model},
  author={Hu, Ke and Hosseini-Asl, Ehsan and Chen, Chen and Casanova, Edresson and Ghosh, Subhankar and {\.Z}elasko, Piotr and Chen, Zhehuai and Li, Jason and Balam, Jagadeesh and Ginsburg, Boris},
  journal={arXiv preprint arXiv:2505.15670},
  year={2025}
}

@article{nanocodec,
  title={NanoCodec: Towards High-Quality Ultra Fast Speech LLM Inference},
  author={Casanova, Edresson and Neekhara, Paarth and Langman, Ryan and Hussain, Shehzeen and Ghosh, Subhankar and Yang, Xuesong and Juki{\'c}, Ante and Li, Jason and Ginsburg, Boris},
  journal={arXiv preprint arXiv:2508.05835},
  year={2025}
}

@article{ntp,
  title={Language models are few-shot learners},
  author={Brown, Tom and Mann, Benjamin and Ryder, Nick and Subbiah, Melanie and Kaplan, Jared D and Dhariwal, Prafulla and Neelakantan, Arvind and Shyam, Pranav and Sastry, Girish and Askell, Amanda and others},
  journal={Advances in neural information processing systems},
  volume={33},
  pages={1877--1901},
  year={2020}
}

@article{kimi,
  title={Kimi-audio technical report},
  author={Ding, Ding and Ju, Zeqian and Leng, Yichong and Liu, Songxiang and Liu, Tong and Shang, Zeyu and Shen, Kai and Song, Wei and Tan, Xu and Tang, Heyi and others},
  journal={arXiv preprint arXiv:2504.18425},
  year={2025}
}

@article{moshi,
  title={Moshi: a speech-text foundation model for real-time dialogue},
  author={D{\'e}fossez, Alexandre and Mazar{\'e}, Laurent and Orsini, Manu and Royer, Am{\'e}lie and P{\'e}rez, Patrick and J{\'e}gou, Herv{\'e} and Grave, Edouard and Zeghidour, Neil},
  journal={arXiv preprint arXiv:2410.00037},
  year={2024}
}

@article{minmo,
  title={Minmo: A multimodal large language model for seamless voice interaction},
  author={Chen, Qian and Chen, Yafeng and Chen, Yanni and Chen, Mengzhe and Chen, Yingda and Deng, Chong and Du, Zhihao and Gao, Ruize and Gao, Changfeng and Gao, Zhifu and others},
  journal={arXiv preprint arXiv:2501.06282},
  year={2025}
}

@article{fireredchat,
  title={FireRedChat: A Pluggable, Full-Duplex Voice Interaction System with Cascaded and Semi-Cascaded Implementations},
  author={Chen, Junjie and Hu, Yao and Li, Junjie and Li, Kangyue and Liu, Kun and Li, Wenpeng and Li, Xu and Li, Ziyuan and Shen, Feiyu and Tang, Xu and others},
  journal={arXiv preprint arXiv:2509.06502},
  year={2025}
}

@article{salmonn,
  title={Salmonn-omni: A codec-free llm for full-duplex speech understanding and generation},
  author={Yu, Wenyi and Wang, Siyin and Yang, Xiaoyu and Chen, Xianzhao and Tian, Xiaohai and Zhang, Jun and Sun, Guangzhi and Lu, Lu and Wang, Yuxuan and Zhang, Chao},
  journal={arXiv preprint arXiv:2411.18138},
  year={2024}
}

@article{loop,
  title={Learning to break the loop: Analyzing and mitigating repetitions for neural text generation},
  author={Xu, Jin and Liu, Xiaojiang and Yan, Jianhao and Cai, Deng and Li, Huayang and Li, Jian},
  journal={Advances in Neural Information Processing Systems},
  volume={35},
  pages={3082--3095},
  year={2022}
}

@article{act-r,
  title={ACT-R: A cognitive architecture for modeling cognition},
  author={Ritter, Frank E and Tehranchi, Farnaz and Oury, Jacob D},
  journal={Wiley Interdisciplinary Reviews: Cognitive Science},
  volume={10},
  number={3},
  pages={e1488},
  year={2019},
  publisher={Wiley Online Library}
}

@misc{qwen2.5,
    title = {Qwen2.5: A Party of Foundation Models},
    url = {https://qwenlm.github.io/blog/qwen2.5/},
    author = {Qwen Team},
    month = {September},
    year = {2024}
}

@article{nemo,
  title={Nemo: a toolkit for building ai applications using neural modules},
  author={Kuchaiev, Oleksii and Li, Jason and Nguyen, Huyen and Hrinchuk, Oleksii and Leary, Ryan and Ginsburg, Boris and Kriman, Samuel and Beliaev, Stanislav and Lavrukhin, Vitaly and Cook, Jack and others},
  journal={arXiv preprint arXiv:1909.09577},
  year={2019}
}

@article{spokenwoz,
  title={Spokenwoz: A large-scale speech-text benchmark for spoken task-oriented dialogue agents},
  author={Si, Shuzheng and Ma, Wentao and Gao, Haoyu and Wu, Yuchuan and Lin, Ting-En and Dai, Yinpei and Li, Hangyu and Yan, Rui and Huang, Fei and Li, Yongbin},
  journal={Advances in Neural Information Processing Systems},
  volume={36},
  pages={39088--39118},
  year={2023}
}

@article{urobench,
  title={Uro-bench: A comprehensive benchmark for end-to-end spoken dialogue models},
  author={Yan, Ruiqi and Li, Xiquan and Chen, Wenxi and Niu, Zhikang and Yang, Chen and Ma, Ziyang and Yu, Kai and Chen, Xie},
  journal={arXiv preprint arXiv:2502.17810},
  year={2025}
}

@article{llamaq,
  title={Spoken question answering and speech continuation using spectrogram-powered llm},
  author={Nachmani, Eliya and Levkovitch, Alon and Hirsch, Roy and Salazar, Julian and Asawaroengchai, Chulayuth and Mariooryad, Soroosh and Rivlin, Ehud and Skerry-Ryan, RJ and Ramanovich, Michelle Tadmor},
  journal={arXiv preprint arXiv:2305.15255},
  year={2023}
}

@article{stepaudio,
  title={Step-audio: Unified understanding and generation in intelligent speech interaction},
  author={Huang, Ailin and Wu, Boyong and Wang, Bruce and Yan, Chao and Hu, Chen and Feng, Chengli and Tian, Fei and Shen, Feiyu and Li, Jingbei and Chen, Mingrui and others},
  journal={arXiv preprint arXiv:2502.11946},
  year={2025}
}

@article{slm_survey1,
  title={Recent advances in speech language models: A survey},
  author={Cui, Wenqian and Yu, Dianzhi and Jiao, Xiaoqi and Meng, Ziqiao and Zhang, Guangyan and Wang, Qichao and Guo, Yiwen and King, Irwin},
  journal={arXiv preprint arXiv:2410.03751},
  year={2024}
}

@article{gen_sdlm,
  title={Generative spoken dialogue language modeling},
  author={Nguyen, Tu Anh and Kharitonov, Eugene and Copet, Jade and Adi, Yossi and Hsu, Wei-Ning and Elkahky, Ali and Tomasello, Paden and Algayres, Robin and Sagot, Benoit and Mohamed, Abdelrahman and others},
  journal={Transactions of the Association for Computational Linguistics},
  volume={11},
  pages={250--266},
  year={2023},
  publisher={MIT Press One Broadway, 12th Floor, Cambridge, Massachusetts 02142, USA~…}
}

@article{turn_based,
  title={Beyond turn-based interfaces: Synchronous llms as full-duplex dialogue agents},
  author={Veluri, Bandhav and Peloquin, Benjamin N and Yu, Bokai and Gong, Hongyu and Gollakota, Shyamnath},
  journal={arXiv preprint arXiv:2409.15594},
  year={2024}
}

@article{FlexDuo,
  title={FlexDuo: A Pluggable System for Enabling Full-Duplex Capabilities in Speech Dialogue Systems},
  author={Liao, Borui and Xu, Yulong and Ou, Jiao and Yang, Kaiyuan and Jian, Weihua and Wan, Pengfei and Zhang, Di},
  journal={arXiv preprint arXiv:2502.13472},
  year={2025}
}

@inproceedings{Duplex_conversation,
  title={Duplex conversation: Towards human-like interaction in spoken dialogue systems},
  author={Lin, Ting-En and Wu, Yuchuan and Huang, Fei and Si, Luo and Sun, Jian and Li, Yongbin},
  booktitle={Proceedings of the 28th ACM SIGKDD Conference on Knowledge Discovery and Data Mining},
  pages={3299--3308},
  year={2022}
}

@inproceedings{Penalty_Decoding,
  title={Penalty Decoding: Well Suppress the Self-Reinforcement Effect in Open-Ended Text Generation},
  author={Zhu, Wenhong and Hao, Hongkun and Wang, Rui},
  booktitle={Proceedings of the 2023 Conference on Empirical Methods in Natural Language Processing},
  pages={1218--1228},
  year={2023}
}

@inproceedings{learn_bias_repetition,
  title={Mitigating the Learning Bias towards Repetition by Self-Contrastive Training for Open-Ended Generation},
  author={Guan, Jian and Huang, Minlie},
  booktitle={Findings of the Association for Computational Linguistics: ACL 2023},
  pages={6897--6909},
  year={2023}
}

@inproceedings{listen_speaking,
  title={Language model can listen while speaking},
  author={Ma, Ziyang and Song, Yakun and Du, Chenpeng and Cong, Jian and Chen, Zhuo and Wang, Yuping and Wang, Yuxuan and Chen, Xie},
  booktitle={Proceedings of the AAAI Conference on Artificial Intelligence},
  volume={39},
  number={23},
  pages={24831--24839},
  year={2025}
}

@inproceedings{Incremental_Processing,
  title={Investigating the Impact of Incremental Processing and Voice Activity Projection on Spoken Dialogue Systems},
  author={Chiba, Yuya and Higashinaka, Ryuichiro},
  booktitle={Proceedings of the 31st International Conference on Computational Linguistics},
  pages={3687--3696},
  year={2025}
}

@inproceedings{vad_analysis,
  title={Analysis of Voice Activity Detection Errors in API-based Streaming ASR for Human-Robot Dialogue},
  author={Yamamoto, Kenta and Takeda, Ryu and Komatani, Kazunori},
  booktitle={Proceedings of the 15th International Workshop on Spoken Dialogue Systems Technology},
  pages={245--253},
  year={2025}
}

@inproceedings{streaming_asr,
  title={Streaming automatic speech recognition with the transformer model},
  author={Moritz, Niko and Hori, Takaaki and Le, Jonathan},
  booktitle={ICASSP 2020-2020 IEEE International Conference on Acoustics, Speech and Signal Processing (ICASSP)},
  pages={6074--6078},
  year={2020},
  organization={IEEE}
}

@inproceedings{rl_sdlm,
  title={Reinforcement Learning Enhanced Full-Duplex Spoken Dialogue Language Models for Conversational Interactions},
  author={Chen, Chen and Hu, Ke and Yang, Chao-Han Huck and Pasad, Ankita and Casanova, Edresson and Wang, Weiqing and Fu, Szu-Wei and Li, Jason and Chen, Zhehuai and Balam, Jagadeesh and others},
  booktitle={Second Conference on Language Modeling}
}

@article{cot,
  title={Chain-of-thought prompting elicits reasoning in large language models},
  author={Wei, Jason and Wang, Xuezhi and Schuurmans, Dale and Bosma, Maarten and Xia, Fei and Chi, Ed and Le, Quoc V and Zhou, Denny and others},
  journal={Advances in neural information processing systems},
  volume={35},
  pages={24824--24837},
  year={2022}
}

@inproceedings{
selfconsistency,
title={Self-Consistency Improves Chain of Thought Reasoning in Language Models},
author={Xuezhi Wang and Jason Wei and Dale Schuurmans and Quoc V Le and Ed H. Chi and Sharan Narang and Aakanksha Chowdhery and Denny Zhou},
booktitle={The Eleventh International Conference on Learning Representations },
year={2023},
url={https://openreview.net/forum?id=1PL1NIMMrw}
}

@article{tree_of_thought,
  title={Tree of thoughts: Deliberate problem solving with large language models},
  author={Yao, Shunyu and Yu, Dian and Zhao, Jeffrey and Shafran, Izhak and Griffiths, Tom and Cao, Yuan and Narasimhan, Karthik},
  journal={Advances in neural information processing systems},
  volume={36},
  pages={11809--11822},
  year={2023}
}

@inproceedings{pal,
  title={Pal: Program-aided language models},
  author={Gao, Luyu and Madaan, Aman and Zhou, Shuyan and Alon, Uri and Liu, Pengfei and Yang, Yiming and Callan, Jamie and Neubig, Graham},
  booktitle={International Conference on Machine Learning},
  pages={10764--10799},
  year={2023},
  organization={PMLR}
}

@article{toolformer,
  title={Toolformer: Language models can teach themselves to use tools},
  author={Schick, Timo and Dwivedi-Yu, Jane and Dess{\`\i}, Roberto and Raileanu, Roberta and Lomeli, Maria and Hambro, Eric and Zettlemoyer, Luke and Cancedda, Nicola and Scialom, Thomas},
  journal={Advances in Neural Information Processing Systems},
  volume={36},
  pages={68539--68551},
  year={2023}
}

@inproceedings{prefix-to-prefix,
  title={STACL: Simultaneous Translation with Implicit Anticipation and Controllable Latency using Prefix-to-Prefix Framework},
  author={Ma, Mingbo and Huang, Liang and Xiong, Hao and Zheng, Renjie and Liu, Kaibo and Zheng, Baigong and Zhang, Chuanqiang and He, Zhongjun and Liu, Hairong and Li, Xing and others},
  booktitle={Proceedings of the 57th Annual Meeting of the Association for Computational Linguistics},
  pages={3025--3036},
  year={2019}
}

@inproceedings{
Monotonic,
title={Monotonic Multihead Attention},
author={Xutai Ma and Juan Miguel Pino and James Cross and Liezl Puzon and Jiatao Gu},
booktitle={International Conference on Learning Representations},
year={2020},
url={https://openreview.net/forum?id=Hyg96gBKPS}
}

@inproceedings{incremental_decoding,
  title={Incremental Decoding and Training Methods for Simultaneous Translation in Neural Machine Translation},
  author={Dalvi, Fahim and Durrani, Nadir and Sajjad, Hassan and Vogel, Stephan},
  booktitle={Proceedings of the 2018 Conference of the North American Chapter of the Association for Computational Linguistics: Human Language Technologies, Volume 2 (Short Papers)},
  pages={493--499},
  year={2018}
}

@inproceedings{
StreamingLLM,
title={Efficient Streaming Language Models with Attention Sinks},
author={Guangxuan Xiao and Yuandong Tian and Beidi Chen and Song Han and Mike Lewis},
booktitle={The Twelfth International Conference on Learning Representations},
year={2024},
url={https://openreview.net/forum?id=NG7sS51zVF}
}

@inproceedings{Medusa,
  title={Medusa: Simple LLM Inference Acceleration Framework with Multiple Decoding Heads},
  author={Cai, Tianle and Li, Yuhong and Geng, Zhengyang and Peng, Hongwu and Lee, Jason D and Chen, Deming and Dao, Tri},
  booktitle={International Conference on Machine Learning},
  pages={5209--5235},
  year={2024},
  organization={PMLR}
}

@inproceedings{Monotonic2,
  title={Monotonic Infinite Lookback Attention for Simultaneous Machine Translation},
  author={Arivazhagan, Naveen and Cherry, Colin and Macherey, Wolfgang and Chiu, Chung-Cheng and Yavuz, Semih and Pang, Ruoming and Li, Wei and Raffel, Colin},
  booktitle={Proceedings of the 57th Annual Meeting of the Association for Computational Linguistics},
  pages={1313--1323},
  year={2019}
}

@inproceedings{he2019streaming,
  title={Streaming end-to-end speech recognition for mobile devices},
  author={He, Yanzhang and Sainath, Tara N and Prabhavalkar, Rohit and McGraw, Ian and Alvarez, Raziel and Zhao, Ding and Rybach, David and Kannan, Anjuli and Wu, Yonghui and Pang, Ruoming and others},
  booktitle={ICASSP 2019-2019 IEEE International Conference on Acoustics, Speech and Signal Processing (ICASSP)},
  pages={6381--6385},
  year={2019},
  organization={IEEE}
}

@inproceedings{skerry2018towards,
  title={Towards end-to-end prosody transfer for expressive speech synthesis with tacotron},
  author={Skerry-Ryan, RJ and Battenberg, Eric and Xiao, Ying and Wang, Yuxuan and Stanton, Daisy and Shor, Joel and Weiss, Ron and Clark, Rob and Saurous, Rif A},
  booktitle={international conference on machine learning},
  pages={4693--4702},
  year={2018},
  organization={PMLR}
}

@article{schlangen2011general,
  title={A general, abstract model of incremental dialogue processing},
  author={Schlangen, David and Skantze, Gabriel},
  journal={Dialogue \& Discourse},
  volume={2},
  number={1},
  pages={83--111},
  year={2011}
}

@inproceedings{sarikaya2002turn,
  title={Turn-based language modeling for spoken dialog systems},
  author={Sarikaya, Ruhi and Gao, Yuqing and Erdogan, Hakan and Picheny, Michael},
  booktitle={2002 IEEE International Conference on Acoustics, Speech, and Signal Processing},
  volume={1},
  pages={I--781},
  year={2002},
  organization={IEEE}
}

@article{veluri2024beyond,
  title={Beyond turn-based interfaces: Synchronous llms as full-duplex dialogue agents},
  author={Veluri, Bandhav and Peloquin, Benjamin N and Yu, Bokai and Gong, Hongyu and Gollakota, Shyamnath},
  journal={arXiv preprint arXiv:2409.15594},
  year={2024}
}

@article{wang2024full,
  title={A full-duplex speech dialogue scheme based on large language model},
  author={Wang, Peng and Lu, Songshuo and Tang, Yaohua and Yan, Sijie and Xia, Wei and Xiong, Yuanjun},
  journal={Advances in Neural Information Processing Systems},
  volume={37},
  pages={13372--13403},
  year={2024}
}

@article{zhang2025llm,
  title={LLM-Enhanced Dialogue Management for Full-Duplex Spoken Dialogue Systems},
  author={Zhang, Hao and Li, Weiwei and Chen, Rilin and Kothapally, Vinay and Yu, Meng and Yu, Dong},
  journal={arXiv preprint arXiv:2502.14145},
  year={2025}
}

@article{sentence-bert,
  title={Sentence-bert: Sentence embeddings using siamese bert-networks},
  author={Reimers, Nils and Gurevych, Iryna},
  journal={arXiv preprint arXiv:1908.10084},
  year={2019}
}

@inproceedings{khouzaimi2016reinforcement,
  title={Reinforcement Learning for Turn-Taking Management in Incremental Spoken Dialogue Systems.},
  author={Khouzaimi, Hatim and Laroche, Romain and Lef{\`e}vre, Fabrice},
  booktitle={IJCAI},
  pages={2831--2837},
  year={2016}
}

@inproceedings{calimeri2021stream,
  title={Stream Reasoning with Incremental Grounding},
  author={Calimeri, Francesco and Ianni, Giovambattista and Pacenza, Francesco and Perri, Simona and Zangari, Jessica},
  booktitle={5th Stream Reasoning Workshop},
  year={2021}
}

@article{glm4voice,
  title={Glm-4-voice: Towards intelligent and human-like end-to-end spoken chatbot},
  author={Zeng, Aohan and Du, Zhengxiao and Liu, Mingdao and Wang, Kedong and Jiang, Shengmin and Zhao, Lei and Dong, Yuxiao and Tang, Jie},
  journal={arXiv preprint arXiv:2412.02612},
  year={2024}
}

@inproceedings{turntakingability,
  title={Reinforcement Learning Enhanced Full-Duplex Spoken Dialogue Language Models for Conversational Interactions},
  author={Chen, Chen and Hu, Ke and Yang, Chao-Han Huck and Pasad, Ankita and Casanova, Edresson and Wang, Weiqing and Fu, Szu-Wei and Li, Jason and Chen, Zhehuai and Balam, Jagadeesh and others},
  booktitle={Second Conference on Language Modeling}
}

@inproceedings{fisher,
  title={The Fisher corpus: A resource for the next generations of speech-to-text.},
  author={Cieri, Christopher and Miller, David and Walker, Kevin},
  booktitle={LREC},
  volume={4},
  pages={69--71},
  year={2004}
}

@inproceedings{fsq,
  title={Low frame-rate speech codec: a codec designed for fast high-quality speech LLM training and inference},
  author={Casanova, Edresson and Langman, Ryan and Neekhara, Paarth and Hussain, Shehzeen and Li, Jason and Ghosh, Subhankar and Juki{\'c}, Ante and Lee, Sang-gil},
  booktitle={ICASSP 2025-2025 IEEE International Conference on Acoustics, Speech and Signal Processing (ICASSP)},
  pages={1--5},
  year={2025},
  organization={IEEE}
}

@misc{whisper,
  doi = {10.48550/ARXIV.2212.04356},
  url = {https://arxiv.org/abs/2212.04356},
  author = {Radford, Alec and Kim, Jong Wook and Xu, Tao and Brockman, Greg and McLeavey, Christine and Sutskever, Ilya},
  title = {Robust Speech Recognition via Large-Scale Weak Supervision},
  publisher = {arXiv},
  year = {2022},
  copyright = {arXiv.org perpetual, non-exclusive license}
}

@inproceedings{bleu,
  title={Bleu: a method for automatic evaluation of machine translation},
  author={Papineni, Kishore and Roukos, Salim and Ward, Todd and Zhu, Wei-Jing},
  booktitle={Proceedings of the 40th annual meeting of the Association for Computational Linguistics},
  pages={311--318},
  year={2002}
}

@inproceedings{webquestions,
  title={Semantic parsing on freebase from question-answer pairs},
  author={Berant, Jonathan and Chou, Andrew and Frostig, Roy and Liang, Percy},
  booktitle={Proceedings of the 2013 conference on empirical methods in natural language processing},
  pages={1533--1544},
  year={2013}
}

@article{abtest,
  title={Subjective Assessment of Quality of Audio and Video Signals by Means of AB Test},
  author={Brachmanski, Stefan}
}
\bibliographystyle{iclr2026_conference}

\appendix
\section{Appendix}
\subsection{Discussion about streaming ASR and chronological thinking}
Using streaming ASR results as thinking content is an intuitive implementation approach to enable SDLMs with the thinking-while-listening ability \citep{streaming_asr, he2019streaming, vad_analysis}. Under the task scenarios and experimental setup of this paper, the performance upper bound of this streaming ASR method corresponds to the performance of the text LLM when fed with the ground-truth user inquiry text. This output reflects the LLM's response without any thinking process; when the task requires reasoning, its performance is significantly affected. In contrast, the chronological thinking method proposed in this paper, which includes analysis of user intent, contextual entity binding, logical reasoning, prerequisite knowledge extraction, and prediction of agent actions, achieves superior reasoning performance. In fact, the comparison between the GT-LM results and our method in Table \ref{tab:gpt} demonstrates this conclusion. Additionally, using streaming ASR results as thinking content is also affected by ASR accuracy. In experiments, we observe that streaming ASR models often insert padding tokens between adjacent text tokens, which disrupts semantic consistency and degrades SDLM's performance.

\label{app:asr}
\subsection{Example of chronological thinking chains}
\label{example}
\begin{figure}[t]
\begin{center}
\begin{minipage}[b]{0.9\linewidth}
  \centering
  \centerline{\includegraphics[width=14cm]{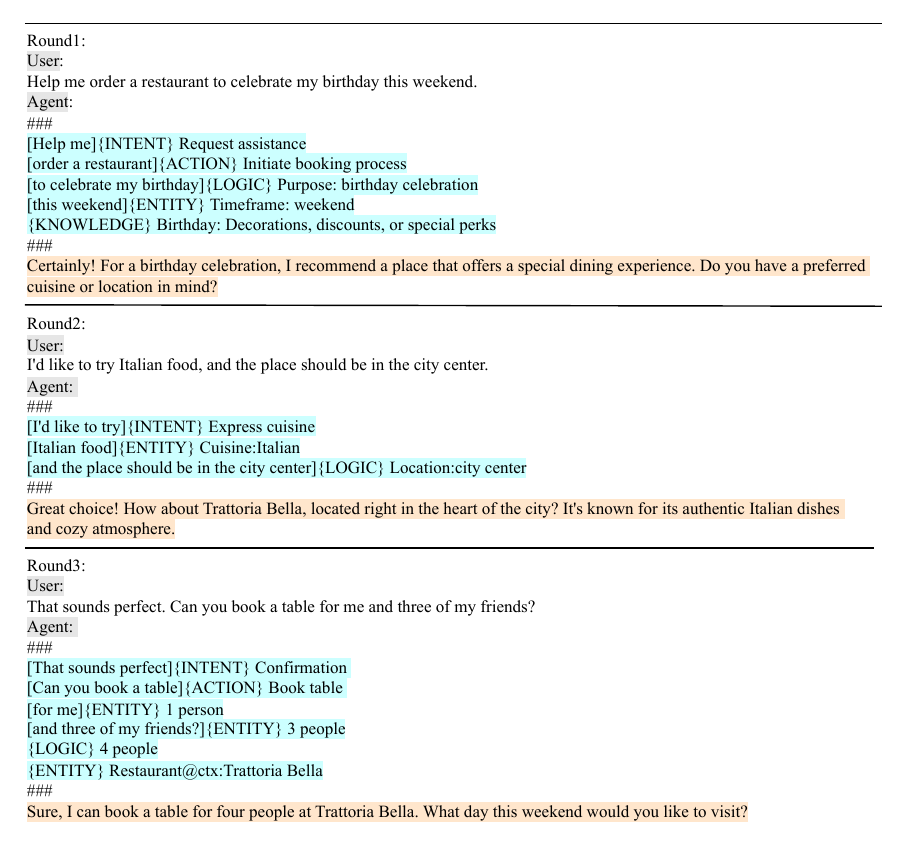}}
\end{minipage}
\caption{A dialogue example with chronological thinking chains.}
\label{fig:example}
\end{center}
\end{figure}

Figure \ref{fig:example} displays a three-turn dialogue example incorporating chronological thinking chains. It includes simple semantic reasoning (inferring the preferred restaurant for a birthday celebration) and mathematical reasoning (me + 3 people = 4 people), as well as cross-turn entity tracking (Trattoria Bella). In the design of the thinking chain, we use ``@ctx:ID" to bind and track entities across dialogue turns, with this functionality implemented within the ENTITY nodes. The usage of ``@ctx:ID" is demonstrated in the last line of the thinking chain in the third dialogue turn.

\begin{wraptable}{r}{0.45\textwidth}
\vspace{-35pt}
\caption{The ratio of dialogue turns where the thinking chain tokens being less than the frame count of user utterances.}
\label{tab:completeratio}
\renewcommand\arraystretch{1}
\begin{center}
\begin{tabular}{lccc}
\toprule
\multicolumn{1}{c}{Dataset}  &\multicolumn{1}{c}{Ratio} 
\\ \midrule
GenConv         & 98.91\% \\
SpokenWOZ-G         & 96.78\% \\
Llamaq-G         & 94.69\% \\
\bottomrule
\end{tabular}
\end{center}
\vspace{-15pt}
\end{wraptable}

\subsection{Completeness of Thinking Chains}
\label{sec:completeness}
Although in Section \ref{thinkingchain} we employ truncation to ensure that thinking tokens do not occupy the original response tokens, the proposed method significantly shortens the length of the generated thinking content by replacing natural language with structured thinking chain nodes. This maximizes the completeness of the thinking process within a given time duration. To verify the completeness of the generated thinking chains during training process, we statistically analyze the frame count of each user utterance and the corresponding token count of the thinking chains, across three training datasets. We then calculate the ratio of dialogue turns where the thinking chain tokens being less than the frame count of user utterances, as summarized in Table \ref{tab:completeratio}. The results demonstrate that for the majority of cases across all three datasets, the token count of the thinking chains remains lower than the frame count of user utterances, indicating that most thinking chains are fully preserved.

\subsection{Subjective Experiments}
\label{sec:abtest}
To ensure a fair comparison, we conduct a blind A/B test where human evaluators are presented with paired responses from the full-duplex SDLM with and without chronological thinking in a random order. Evaluators are asked to select the preferred response based on audio fidelity and response content quality, or mark them as a tie if no significant difference is observed.

We recruit 10 fluent English speakers as evaluators, where each participant assesses 20 audio samples selected randomly from the test dataset. These subjects are either from English-speaking countries or possess over seven years of speaking experience, ensuring high proficiency. To ensure precise evaluation, evaluators are allowed to replay the audio repeatedly but are required to listen to each sample at least three times before providing a rating.

\subsection{LLM USAGE STATEMENT}
We used ChatGPT only for minor language editing to improve clarity and conciseness. No part of the research idea, methodology, or analysis was generated by LLMs.

\end{document}